\documentclass[letterpaper, 10 pt, conference]{ieeeconf}  

\IEEEoverridecommandlockouts                              

\usepackage{cite}
\usepackage{amsmath,amssymb,amsfonts}
\usepackage{graphicx}
\usepackage{svg}
\usepackage{textcomp}
\usepackage{xcolor}
\usepackage{algorithm}
\usepackage{algpseudocode}
\usepackage{glossaries}
\usepackage{subcaption}
\usepackage{float}

\title{\LARGE \bf
Efficient Online Learning and Adaptive Planning for Robotic Information Gathering Based on Streaming Data
}

\author{Sanjeev Ramkumar Sudha, Joel Jose and Erlend M. Coates
\thanks{The authors are with the Department of ICT and Natural Sciences, Norwegian University of Science and Technology (NTNU), Ålesund, Norway.
        {\tt\small \{sanjeev.k.r.sudha,joel.j.chirapanath,
        erlend.coates\}@ntnu.no}}%
}

\begin{document}

\maketitle
\thispagestyle{empty}
\pagestyle{empty}

\begin{abstract}

Robotic information gathering (RIG) techniques refer to methods where mobile robots are used to acquire data about the physical environment with a suite of sensors. Informative planning is an important part of RIG where the goal is to find sequences of actions or paths that maximize efficiency or the quality of information collected. Many existing solutions solve this problem by assuming that the environment is known in advance. However, real environments could be unknown or time-varying, and adaptive informative planning remains an active area of research. Adaptive planning and incremental online mapping are required for mapping initially unknown or varying spatial fields. Gaussian process (GP) regression is a widely used technique in RIG for mapping continuous spatial fields. However, it falls short in many applications as its real-time performance does not scale well to large datasets. To address these challenges, this paper proposes an efficient adaptive informative planning approach for mapping continuous scalar fields with GPs with streaming sparse GPs. Simulation experiments are performed with a synthetic dataset and compared against existing benchmarks. Finally, it is also verified with a real-world dataset to further validate the efficacy of the proposed method. Results show that our method achieves similar mapping accuracy to the baselines while reducing computational complexity for longer missions.



\end{abstract}


\section{Introduction}

Mobile robots are increasingly used for information gathering as they can operate in conditions that could otherwise be dangerous and inaccessible to humans. 
Environmental monitoring refers to observational techniques used in gathering information about a physical quantity of significance at a specific location. Ideally, a number of static sensors could be placed over the area of interest so that continuous measurements are observed. The sensor placement problem is one way to solve this problem by finding the optimal sensor locations to maximize efficiency \cite{krause2008near}. However, using multiple static sensors over large regions is both impractical and very expensive. The use of mobile robots equipped with sensors to collect information is becoming more prevalent \cite{dunbabin2012robots}. Such methods are referred to as robotic information gathering (RIG). Some applications in environmental monitoring include bathymetry mapping \cite{agrawal2024oas, torroba2022fully}, monitoring for contaminants \cite{hitz2017adaptive}, oceanographic surveys \cite{berget2019adaptive}, and terrain mapping \cite{popovic2020informative}. Other applications also include 3D reconstruction \cite{schmid2020efficient}, anomaly detection \cite{blanchard2022informative}, and search and rescue \cite{meera2019obstacle}.

\begin{figure}[!htbp]
    \centering
    \includegraphics[width=0.98\linewidth]{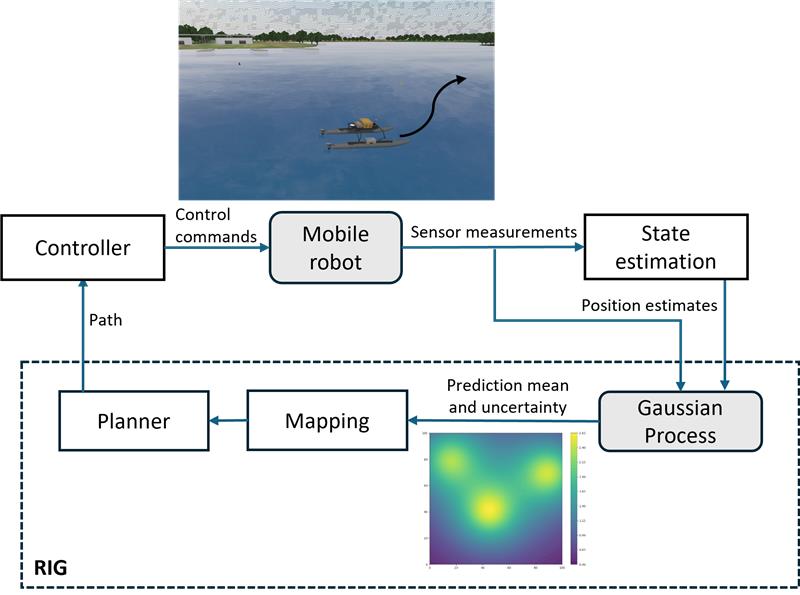}
    \caption{Overview of the overall framework. In this study, we consider active mapping for the reconstruction of continuous spatial fields. A Gaussian process (GP) is used for mapping the scalar field. We utilize streaming sparse GP regression for online learning of the GP hyperparameters. A sampling-based planner is used for replanning paths for the robot.
    }
    \label{fig:flow}
\end{figure}

Planning a safe and feasible path to collect the most informative measurements is an active area of research that is collectively termed informative path planning \cite{Bai2021, popovic2024learning}. There are several challenges associated with this problem and it is known to be NP-hard. These problems are also typically solved as optimization problems with the assumption that the environment is time-invariant and known a priori. However, the real environment with which the robot interacts is unpredictable and dynamic, where the states or the observable information may vary rapidly in real time. Additionally, the environment may only be partially observable, and there might be uncertainty in the measurements as a result. Designing algorithms that balance exploration and exploitation is another aspect to consider. In cases where the prior knowledge about the environment to be mapped is unavailable, adaptive or online planning is required, and the path has to be replanned in real-time using the measurements collected in situ.

Representation of the map is also crucial for active mapping tasks. In online informative planning scenarios, the quality of the predicted paths is dependent on the belief about the environment. Gaussian Processes (GPs) are widely used for mapping continuous spatial fields for information gathering tasks. However, learning the hyperparameters of a GP kernel can be computationally expensive. This computational burden prevents GPs from being used in many real-time applications. Some approximations, such as sparse GPs \cite{snelson2005sparse}, and variational sparse GPs \cite{hensman2013gaussian, titsias2009variational} provide a more scalable alternative. These approximate GPs still need the full dataset to optimize the hyperparameters, which could become intractable for long-term monitoring. In this paper, we use streaming sparse GPs \cite{bui2017streaming} to incrementally update the map with streaming data.


Sampling-based algorithms such as rapidly exploring random trees (RRTs) and probabilistic roadmaps (PRMs) have been used to tackle the informative planning problem with guarantees of asymptotic optimality \cite{Hollinger2014, ghaffari2019sampling, moon2022tigris}. These methods, however, consider planning in an offline fashion, i.e. under the assumption that the environment is known a priori. RRTs were later adopted by \cite{Viseras2019} for adaptive informative planning with online learning of Gaussian processes (GPs). They search for informative stations based on a heuristic and then plan informative paths with RRTs. Monte-Carlo tree search is another method that has been used in adaptive informative path planning in partially observable environments \cite{flaspohler2019information}. Hitz et al \cite{hitz2017adaptive} employed an evolutionary optimization method for informative planning in continuous state spaces. Receding-horizon planners \cite{schmid2020efficient,bircher2016receding} combine principles from optimal control and sampling-based planning algorithms, and have proven to be efficient for 3D exploration. Bayesian optimization (BO) has also been effectively applied to online informative planning \cite{marchant2014bayesian,blanchard2022informative} as it naturally lends itself to GPs.

Chen and Liu \cite{chen2019long} compared approximate GP regression techniques in terms of their effectiveness and accuracy for bathymetry mapping. Chen et al \cite{Chen2024} proposed a novel mapping method for online information gathering with a non-stationary kernel. In \cite{torroba2022fully}, the authors used stochastic variational GPs for localization and bathymetry mapping. Both these studies focus on scalable online mapping but do not discuss adaptive planning in detail. Jakkala and Akella \cite{jakkala2024multi} proposed an informative planning approach with sparse GPs, but do not consider online planning. A sparse GP regression method for learning and online planning was proposed in \cite{Ma2017}. They discretize the environment and solve planning as a routing problem. This approach might not scale well for long term monitoring in bigger environments. Viseras et al \cite{Viseras2019} used sampling-based planners for mapping scalar spatial fields along with online learning of GPs. However, they did not consider the cost of GP training for large datasets, and performed replanning, training the GP, and data acquisition in separate stages.

This study overcomes limitations in existing work by presenting an efficient adaptive planning and online learning framework for mapping spatial fields of interest. An efficient sampling based planning strategy is used for adaptive replanning in continuous spaces. Additionally, our work uses streaming sparse GPs to alleviate the computational cost of retraining Gaussian processes in real time. Therefore, the main contributions of this work are:

\begin{itemize}
    \item An adaptive information gathering framework with efficient replanning for mapping initially unknown spatial fields. With our replanning strategy, replanning and online learning of GPs can be performed in parallel without having to stop to replan, in contrast to previous related work.
    \item A scalable online mapping approach with streaming sparse GPs for learning scalar spatial fields that overcomes limitations of training GPs for online mapping. The reduced computational burden of streaming sparse GPs provides a scalable method for learning maps in long-term monitoring scenarios.
\end{itemize}

The remainder of the paper is organized as follows: Section \ref{sec:2} briefly reviews the relevant theory behind online GP regression that is used in this study. Section \ref{sec:3} describes our informative planning framework. Our method is experimentally evaluated in Section \ref{sec:4}. Finally, Section \ref{sec:5} concludes the study by discussing limitations and future research directions.


\section{Preliminaries}
\label{sec:2}


\subsection{Gaussian Process Regression}

GPs are a widely utilized modelling tool in information gathering and geostatistics, providing an elegant way to predict the mean values of stochastic process variables as well as the associated uncertainty \cite{williams2006gaussian}. A GP is usually represented by its mean $\mu (\mathbf{x})$ and covariance function $K_{\theta}(\boldsymbol{x}, \boldsymbol{x}^{\prime})$:
\begin{equation}
\mathrm{f}({\boldsymbol{x}}) \sim \mathcal{G} \mathcal{P}\left(\mu(\boldsymbol{x}), K_{\boldsymbol{\theta}}\left(\boldsymbol{x}, \boldsymbol{x}^{\prime}\right)\right) ,
\label{eq:gp}
\end{equation} 


\noindent
where $\mathrm{f}(x)$ is the target distribution, $\boldsymbol{\theta}$ denotes the hyperparameters of the covariance function $K_{\theta}(x,x')$, which models the correlation between arbitrary points in the space. To optimize the hyperparameters $\theta$ of the covariance function, the log marginal likelihood $\log p(\mathbf{y} | \theta)$ is maximized as

\begin{equation}
\begin{gathered}
\log p(\mathbf{y} \mid \boldsymbol{x}, \boldsymbol{\theta})=-\frac{1}{2} \mathbf{y}^{\top} K_y^{-1} \mathbf{y}-\frac{1}{2} \log \left|K_y\right|-\frac{n}{2} \log 2 \pi \\
\\
\boldsymbol{\theta}^*=\max _{\boldsymbol{\theta}} \log p(\mathbf{y} \mid \boldsymbol{x}, \boldsymbol{\theta}) ,
\end{gathered}
\end{equation}
where $\boldsymbol{x}$ and $\mathbf{y}$ denote the set of training inputs and targets. $K_y = k_{\theta}(x,x) + \sigma^2 \mathbf{I}$ is the sum of the covariance matrix of the training inputs and the diagonal noise matrix.

\subsection{Sparse Gaussian Processes}

Although exact GP regression provides a closed-form solution, its run time scales to $O(n^3)$ for training and $O(n^2)$ for inference, making it intractable for large datasets. This makes it unsuitable for many applications. Sparse GP regression (SGPR)  offers an alternative, as it uses a set of inducing inputs $Z$ to compute the posterior distribution rather than using the entire set of inputs. 




In variational sparse GPs \cite{titsias2009variational}, a variational distribution $q(\mathbf{u})$, where $\mathbf{u}$ is the set of inducing points, is learned by minimzing the Kullback-Leibler (KL) divergence between the exact posterior and the approximate posterior. As the KL divergence is intractable, the evidence lower bound (ELBO), which is the lower bound of the evidence $p(\mathbf{y}|\boldsymbol{\theta})$, is maximized as shown in the following equation:

\begin{equation}
\begin{aligned}
\mathbb{ELBO}[\boldsymbol{\theta}, q(\mathbf{f})]= & \log p(\mathbf{y}|\boldsymbol{\theta})-\mathrm{KL}\left[q(\mathbf{f}, \mathbf{u}) \| p(\mathbf{f}, \mathbf{u} \mid \mathbf{y})\right] \\
& = \int  q(\mathbf{f}) \log  \frac{p(\mathbf{y}, \mathbf{f} \mid \boldsymbol{\theta})}{q(\mathbf{f})}  d\mathbf{f}
\label{eq:elbo_svgp}
\end{aligned}
\end{equation}

The variational parameters can be learned directly through \eqref{eq:elbo_svgp} by variational inference\cite{hensman2013gaussian}. The ELBO can be collapsed to give an analytical solution for sparse GP regression, in the case of a Gaussian likelihood~\cite{titsias2009variational}:

\begin{equation}
\mathbb{ELBO}=\log \mathcal{N}\left(\mathbf{y} \mid \boldsymbol{\theta}, Q_{y y}\right)-\frac{1}{2 \sigma^2} \operatorname{tr}\left(K_{f f}-Q_{f f}\right)
\label{eq:collapsed_sgpr}
\end{equation}
where $Q_{ff} = K_{uf}^T K_{uu}^{-1} K_{uf}$ and $Q_{yy}=Q_{ff}+\Sigma$. The hyperparameters of the kernel can be learned by optimizing the ELBO w.r.t $\boldsymbol{\theta}$. The training and prediction time in sparse GPs is $O(nm^2)$ and $O(m^2)$ respectively, where $m$ is the number of inducing inputs. If the uncollapsed bound \eqref{eq:elbo_svgp} is used directly to learn the parameters with variational inference, the training runtime is further reduced to $O(bm^2)$ where $b$ is the batch size used to update the ELBO. 

\subsection{Streaming Sparse Gaussian Processes}
\label{sec:ssgp}

Streaming sparse GPs (SSGPs) \cite{bui2017streaming} offers a way to update the model incrementally in a streaming setting, i.e. a new batch of data is only used once to update the model and then forgotten. This makes it particularly suitable for scalable online learning and active informative gathering scenarios where a continuous stream of data is received. The exact posteriors before and after considering the new data, $p\left(\mathbf{f} |\mathbf{y}_{\text{old}}\right)$ and $p\left(\mathbf{f} |\mathbf{y}_{\text{old}}, \mathbf{y}_{\text{new }}\right)$ respectively, are given by

\begin{equation}
\begin{aligned}
& p\left(\mathbf{f} |\mathbf{y}_{\text{old}}\right)=\frac{p\left(\mathbf{f} | \boldsymbol{\theta}_{\text{old}}\right) p\left(\mathbf{y}_{\text{old }} | \mathbf{f}\right)} {\mathcal{Z}_1\left(\boldsymbol{\theta}_{\text{old }}\right)}  \\
& p\left(\mathbf{f} |\mathbf{y}_{\text{old}}, \mathbf{y}_{\text{new }}\right)=\frac{p\left(\mathbf{f} | \boldsymbol{\theta}_{\text{new}}\right) p\left(\mathbf{y}_{\text{old}} | \mathbf{f}\right) p\left(\mathbf{y}_{\text{new}} |\mathbf{f}\right)}{\mathcal{Z}_2\left(\boldsymbol{\theta}_{\text{new}}\right)} ,
\end{aligned}
\end{equation}
where $\mathcal{Z}_1(\boldsymbol{\theta}_{\text{old}})$ and $\mathcal{Z}_2(\boldsymbol{\theta}_\text{new})$ are the normalization parameters. Let $q'(\mathbf{f})$ and $q(\mathbf{f})$ denote the approximate posteriors before and after the new batch of data is used to update the distribution. Eliminating $p(\mathbf{y}_{\text{old}}|\mathbf{f})$ by substitution makes the new posterior independent of the old data as given below in \eqref{eq:ssgp}

\begin{equation}
\hat{p}\left(\mathbf{f}|\mathbf{y}_{{old}}, \mathbf{y}_{\text {new}}\right)=\frac{\mathcal{Z}_1\left(\boldsymbol{\theta_{\text {old}}}\right)}{\mathcal{Z}_2\left(\boldsymbol{\theta_{\text{new}}}\right)} p\left(\mathbf{f}|\boldsymbol{\theta_{\text{new}}}\right) p\left(\mathbf{y}_{\text{new}}| \mathbf{f}\right) \frac{q'(\mathbf{f}
)}{p\left(\mathbf{f}| \boldsymbol{\theta_{\text{old}}}\right)}
\label{eq:ssgp}
\end{equation}
Taking the KL divergence between this new approximate posterior $\hat{p}(\mathbf{f}|\mathbf{y}_{\text{old}})$ and the approximate posterior $q(\mathbf{f})$ leads to the online ELBO given as
\begin{equation}
\begin{aligned}
\mathbb{ELBO} &=  \mathbb{E}_{q(f)} \left[ \log p(\mathbf{y}|\mathbf{f}) \right] - \mathrm{KL}[q(\mathbf{u}) \| p(\mathbf{u})] \\
&+ \mathrm{KL}[q(\mathbf{u'}) \| p(\mathbf{u'})] - \mathrm{KL}[q(\mathbf{u'}) \| q'(\mathbf{u'})],
\label{eq:ssgp_elbo}
\end{aligned}    
\end{equation}
where  $\mathbf{u'}$ is the old set of inducing points, $q'$ is the old posterior, and $p'$ is the old prior. The first two terms are the same as the uncollapsed bound of SGPR, and the last two terms are added KL regularization terms. This online ELBO can be further simplified to a collapsed form, similar to SGPR. For the full derviation, we refer the reader to \cite{bui2017streaming}.




\section{Methodology}
\label{sec:3}

\subsection{Problem Statement}

The objective of this study is to collect information and reconstruct a map of a 2D spatial field in an environment that is a priori unknown, as accurately as possible by minimizing the root mean squared error (RMSE) between the ground truth and the prediction. The robot collects samples along its path and uses a surrogate model to update its belief of the environment. The path to be followed by the robot is decided by an informative planner. A GP model is used to map the physical process in the environment, as described in \ref{sec:ssgp}. The hyperparameters of the GP are updated periodically with batches of new data. The planning is adaptive and utilizes the surrogate model that is updated periodically, to replan and make decisions. The overall framework is depicted in Fig. \ref{fig:flow}. 

Informative planning refers to finding paths that maximize an information-theoretic measure that quantifies the quality of the information collected, while adhering to a certain budget, such as path length or battery constraints. The informative path planning problem can be defined as
\begin{equation}
\mathcal{T}^*={\arg\max} \, I(\mathcal{T}) \text { s.t. } C(\mathcal{T}) \leq \mathcal{B} ,
\label{eq:ipp}
\end{equation}
where $\mathcal{T}^*$ and $\mathcal{T}$ represent the optimal trajectory and an arbitrary trajectory respectively, $C(\mathcal{T})$ is the cost function, and $I(\mathcal{T})$ denotes the information-theoretic measure of an arbitrary trajectory $\mathcal{T}$. $B$ denotes the total budget, that is, the maximum allowed cumulative cost over the trajectory. 
\subsection{Mapping}

For mapping with GPs, we use the radial basis function kernel and initialize the GP with a zero mean, as we assume no prior knowledge about the region of interest to be mapped. As the robot begins collecting more data while sampling, the surrogate model subsequently improves its predictions about the environment. This surrogate model is updated using SSGP regression, as summarized in Section~\ref{sec:ssgp}. We assume a streaming setting, meaning the data is seen only once to update the model and then discarded.

The number of inducing points is progressively increased to maintain it at $15\%$ of the total number of measurements. The set of inducing points is updated by performing pivoted Cholesky decomposition, as optimizing the inducing points with the variational distribution does not perform well in the online setting \cite{Chen2024}. Once the inducing points are updated, the GP is trained on the new batch of streaming data with minibatch gradient descent to learn the hyperparameters. The map is updated at regular intervals with an update frequency, $t = t_{u}$. After an update, the memory of the measurements used to perform the update is cleared.




\subsection{Adaptive planning}
\label{sec:planning}

The approach employs a sampling-based strategy to pick locations for sampling with potentially high information gain, based on the current knowledge about the environment, which is represented by the predictive mean of the GP. These new measurements are then used to update this surrogate model, which is then used by the planner to make decisions about the path to sample next, and this process is repeated iteratively until a termination condition is reached. In this section, details about this planning algorithm are presented. 

An RRT* \cite{karaman2011sampling} is used to build a tree $\mathbb{T}$ with the initial location being the root node. Each node of the tree is a tuple, $\nu = \{(x, y), h(\nu), d(\nu)\}$ where $(x, y)$ is the position of the node, $h(\nu)$ is the cumulative entropy, calculated from the GP posterior prediction. The distance of the node from the root $d(\nu)$ is calculated as $d(\nu) = \texttt{compute\_distance}(\nu, \nu') + d(\nu')$, where $\nu'$ is the parent node. During the expansion, the cumulative entropy $h(\nu)$ is also computed as $h(\nu) = H(\nu) + h(\nu')$, the sum the entropy at current position and the cumulative entropy at the parent node. At the beginning of each iteration, the point $x^{*}$ on the map that greedily maximizes the entropy $H(x)$ is computed with Bayesian optimization. The tree is expanded towards this point $x^{*}$ with a probability of $\epsilon$ and towards a random point with probability $1-\epsilon$. This parameter ensures that the tree expansion is partially exploitative and greedy. The tree is expanded until a certain number of nodes are added and a decision step is reached. At this stage, the path in the tree that lies within a certain distance $r$ from the root node $\nu_{0}$, and maximizes the information measure $I$ is chosen. The informativeness is quantified according to \eqref{eq:entr}. At the next planning iteration, the tree is expanded with the same process, with the new root node being the final node on the previously replanned path. This process is repeated until a termination condition is met. The mission is terminated once the budget is exhausted. A condensed version of this algorithm is provided in Algorithm \ref{algo:framework}.

\begin{algorithm}[!htbp]
  \caption{Online information gathering}
  \label{algo:framework}
  \begin{algorithmic}[1]
\State Collect initial dataset $X_{0}, Y_{0}$ by running a random trajectory
\State Initialize $\mathcal{GP}$ and update with initial data $X_{0}, Y_{0}$
\While{$t < \mathcal{B}$}
\State{$X \gets (x_{t}, y_{t}), Y \gets f(x_{t},y_{t})$}
\If{$t \% t_u ==0$}
\State{Optimize hyperparameters with SSGP regression for new batch $X, Y$}
\State {$X \gets \emptyset, Y \gets \emptyset$}
\EndIf
 \If{replan condition == True}
\State{$\mathcal{P}_{new}$ = \texttt{Replanning()}}
\EndIf
      \EndWhile
    \State \Return Map with $\mathcal{GP}$
  \end{algorithmic}
\end{algorithm}

\begin{algorithm}[!htbp]
  \caption{Replanning}
  \label{algo:replan}
  \begin{algorithmic}[1]
\Ensure Path $\mathcal{P}$
\State{Compute $x^{*} \gets \arg\max_{x} H(x)$ with Bayesian optimization}
\While{$n(\mathbb{T}) < n_{max}$}
 \State Initialize tree with root node, $\mathbb{T} = \{\nu_{0}\}$
    
\State{Extend towards $x^{*}$ with probability $\epsilon$, and towards random point with probability $1 - \epsilon$}
\State{New node $\nu_{i}$ with nearest neighbour $\nu_{near}$}
\If{\texttt{collision\_check}($\nu_{i},\nu_{near}$) == False}
\State{ $ \mathbb{T} \gets \nu_{i}$}
\State{\texttt{rewire($\nu_{i}$)}}
\Else
\State{continue}
\EndIf
     \EndWhile
\State Find $\mathcal{S} = \arg \max_{\nu} I(\nu)$ s.t $\nu \in \mathbb{T}$ and \texttt{compute\_distance}$(\nu_{0},\mathcal{S}) <= r$

    \State \Return $\mathcal{P}$
      \State $\mathcal{P} \gets \texttt{extract\_path}(\nu_{0},\mathcal{S})$
\State $\nu_{0} \gets \mathcal{S}$
  \end{algorithmic}
\end{algorithm}


Mutual information is a commonly used information-theoretic measure in informative planning. However, it is computationally expensive to calculate and therefore impractical for an online planner that needs to run in real-time with limited resources on a mobile robot, as shown in \cite{Viseras2019}. Therefore, entropy is used as the information gain function in this study. To evaluate the quality of a path, the average of the entropy along a path in the tree is calculated. The information gain at a node $I(\nu)$, is defined as the average of the entropy at the nodes along a path, defined as

\begin{equation}
\begin{aligned}
    I(\nu_{k}) = \frac{1}{d(\nu_{r},\nu_{k})}\sum_{i=\nu_{0}}^{i=\nu_{k}} H(i) \\
    H(\nu) =  \frac{1}{2} log \Bigl(2\pi e\sigma_{\nu}^2\Bigr)   , 
\end{aligned}
\label{eq:entr}
\end{equation}
where $\sigma_{x}$ is the posterior standard deviation of the GP at $x$, and $H(\nu)$ is the entropy at a location $\nu$. The information gain $I(\nu_k)$ at an arbitrary node $\nu_k$ is the the sum of the entropies along its path from the root node $\nu_{1}$ weighted by the length of the path or cost $d(\nu_r,\nu_k)$. Using average entropy as the information measure leads to paths that reduce uncertainty or equivalently reducing the RMSE in the prediction of the spatial field.

\begin{figure*}[h!]
\centering
    \begin{subfigure}{0.31\textwidth}
    \centering
    \includegraphics[width=0.98\linewidth]{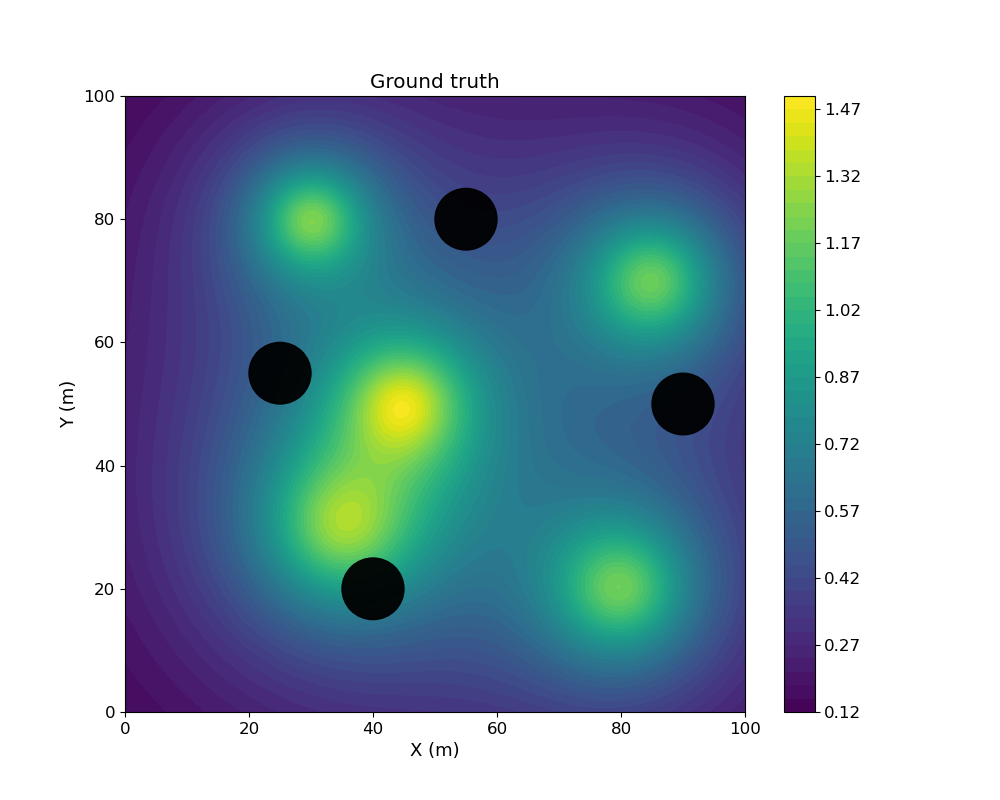}
    \caption{Ground truth values.}
    \label{fig:groundtruth}
    \end{subfigure}
    \begin{subfigure}{0.33\textwidth}
        \centering
        \includegraphics[width=0.98\linewidth]{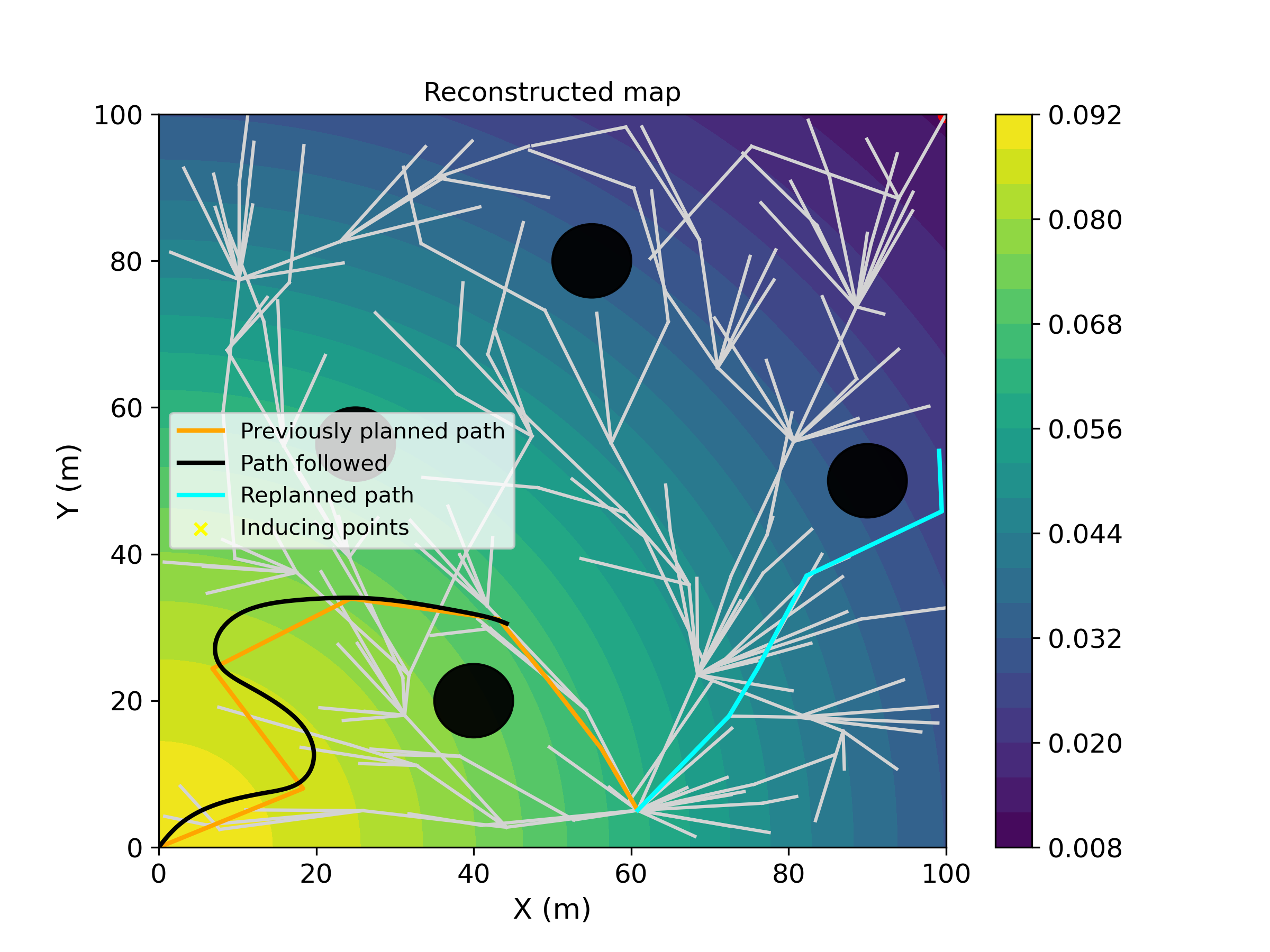}
        \caption{Predicted mean after $t = 100$s.}
        \label{fig:example1}
    \end{subfigure}
    \begin{subfigure}{0.33\textwidth}
        \centering
        \includegraphics[width=0.98\linewidth]{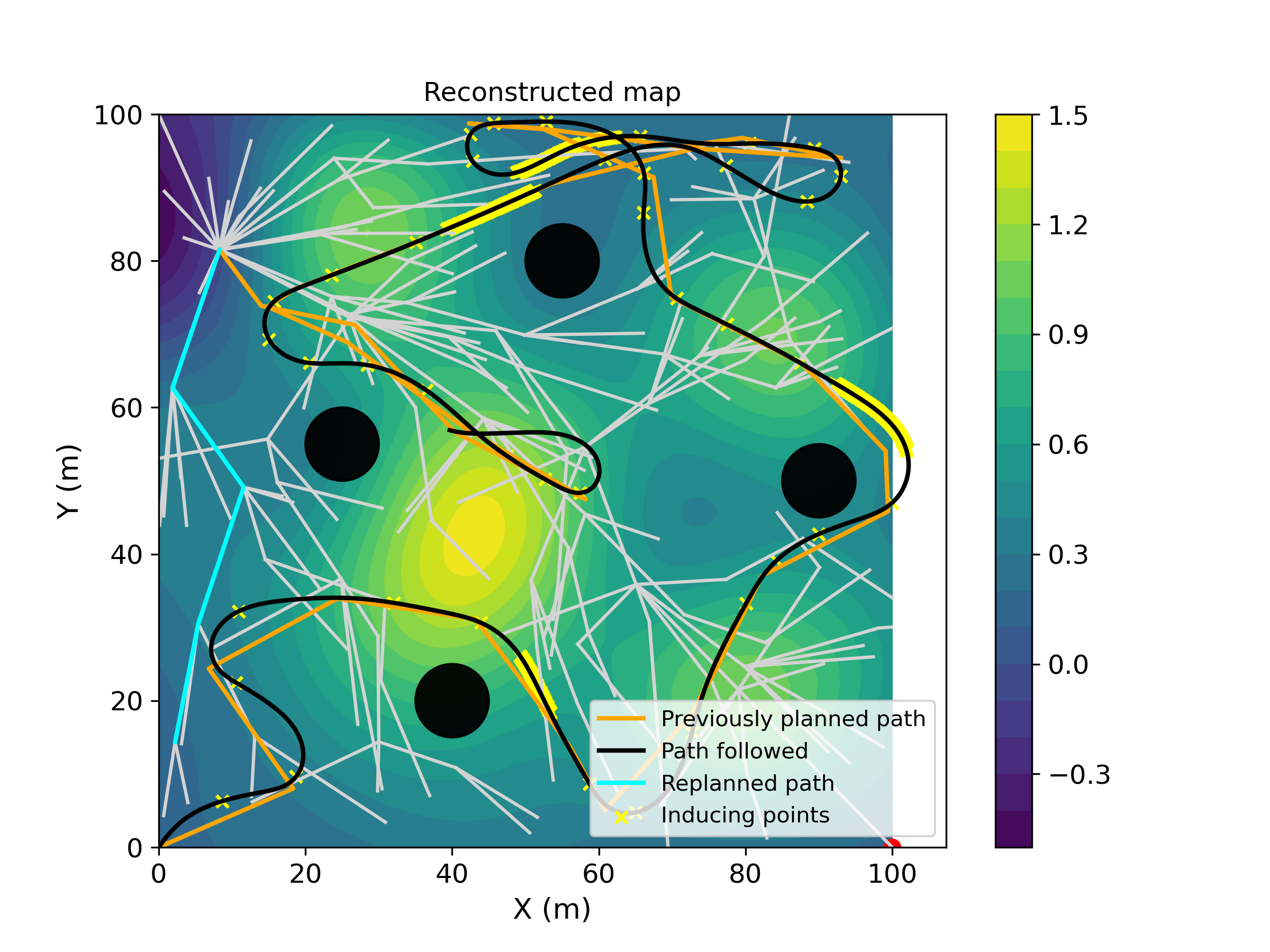}
        \caption{Predicted mean after $t = 400$s.}
        \label{fig:example2}
    \end{subfigure}
    \caption{Example result showing the ground truth values, and the reconstructed map at $t =100$ and $t=400$s. The robot starts from the origin with no initial knowledge of the map, i.e. GP with zero mean.  Fig. \ref{fig:example1} shows the planned trajectory, path followed, and the reconstructed map after $t=100$s. Fig. \ref{fig:example2} depicts the same after $t=400$s. The black regions denote obstacles. }
    \label{fig:illustrate}
\end{figure*}

To ensure that the mapping and replanning stages run parallel realtime, we perform replanning on the fly while the robot is collecting measurements along previously planned paths. Our replanning condition just calculates the vehicle's current position along the current path. The replanning procedure is described in Alg \ref{algo:replan}. An illustration of this replanning procedure, along with the predictions, is depicted in Fig. \ref{fig:illustrate}.

\section{Results and Discussion}
\label{sec:4}


\subsection{Simulation results}

First, we compare our adaptive planning strategy to other baselines for reconstruction of a synthetic map of size $100 \times 100$ meters containing obstacles, as shown in Fig.~\ref{fig:groundtruth}. To verify our algorithm in a real-time monitoring scenario with streaming sensor data, experiments are performed with an uncrewed surface vessel (USV) in the VRX-simulator \cite{bingham2019toward} in Gazebo. We assume that the USV is equipped with a point sensor that publishes data from the synthetic map at 5 Hz. The surge speed of the vehicle is controlled to be 1 m/s. Line of sight guidance \cite{lekkas2013line} is used to generate smooth and continuous trajectories for the vehicle. All experiments are conducted on a computer equipped with an Intel Core i7-12700H processor, 32 GB DDR5 RAM, and an NVIDIA GeForce RTX 3050 Ti GPU. 

Two planning baselines are selected for evaluating our method as described in Alg \ref{algo:framework}. The RIG-tree \cite{Hollinger2014} is a state-of-the-art informative planning algorithm. It is used in a receding horizon way, alternating between replanning and data acquisition stages, similar to the implementation in \cite{Viseras2019}. Another baseline is a random planner that generates random collision-free paths. The map for the random planner is updated at constant intervals. For mapping, we compare between exact GP regression, variational sparse GP regression, and SSGP regression. Ten trials are run for these mapping and planning strategies, and the metrics are reported in Table \ref{tab:table1}. 

\begin{figure}[h!]
\centering
    \begin{subfigure}{0.48\textwidth}
    \centering
    \includegraphics[width=0.85\linewidth]{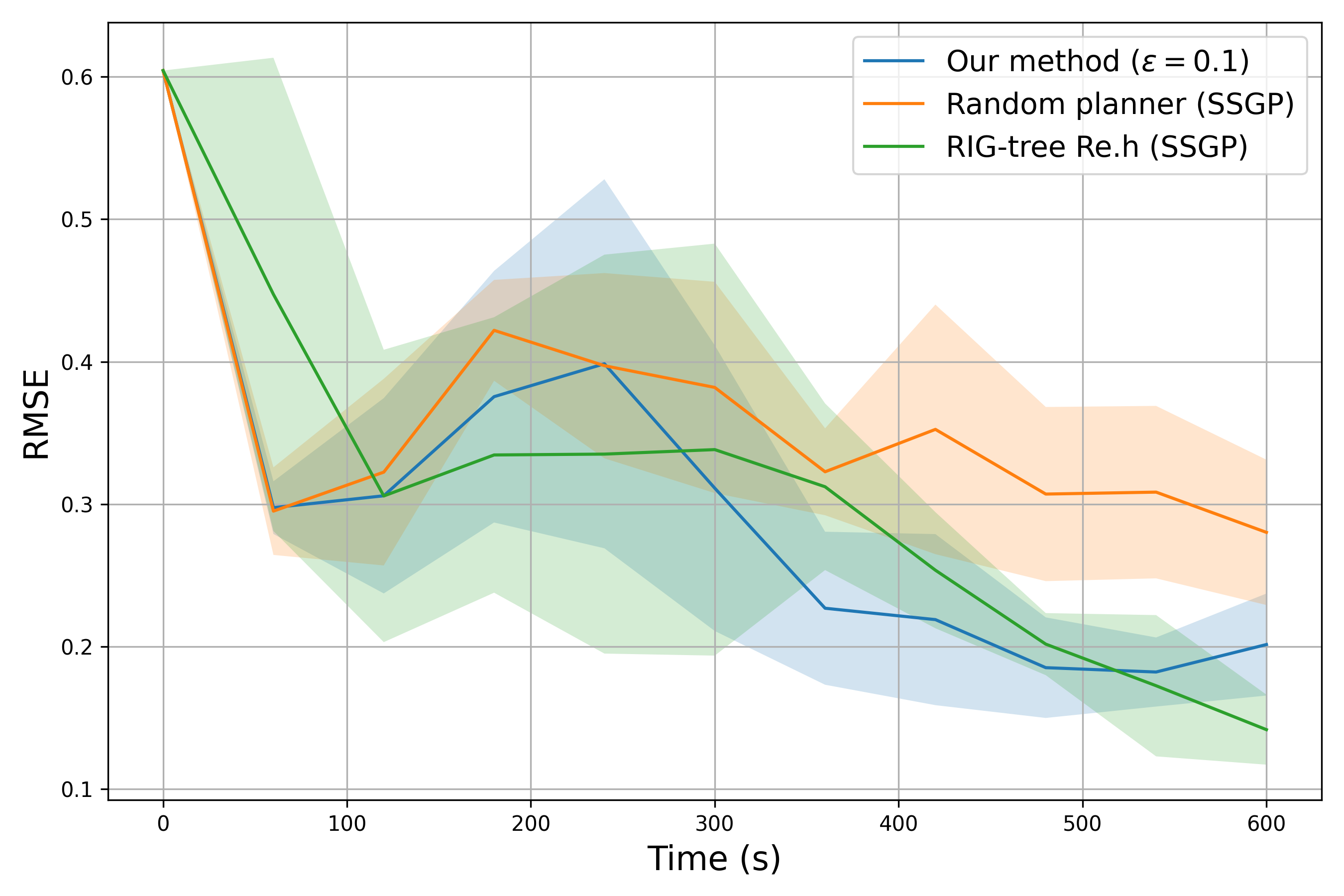}
    \caption{RMSE}
    \label{fig:comp1}
    \end{subfigure}
    \begin{subfigure}{0.48\textwidth}
        \centering
        \includegraphics[width=0.85\linewidth]{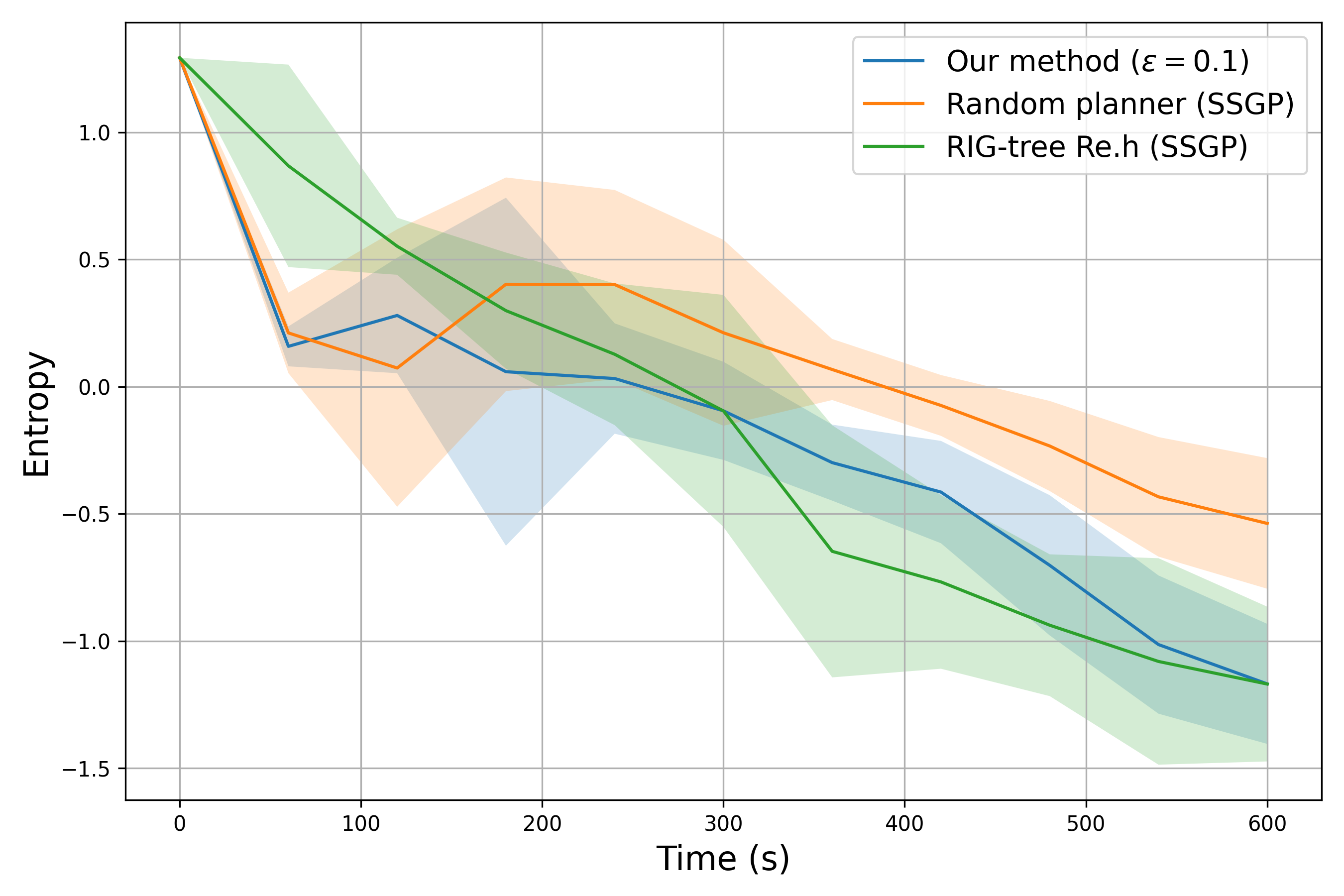}
        \caption{Entropy}
        \label{fig:comp2}
    \end{subfigure}
    \caption{RMSE and Entropy with different planners on the synthetic dataset.}
    \label{fig:comp_planners}
\end{figure}

It can be observed from Table \ref{tab:table1} that our method is as accurate as the RIG-tree method in mapping the environment in terms of the RMSE. It is also observed that our method traverses longer distances than the RIG-tree, due to our replanning strategy that replans paths on the go. SSGP regression is also as accurate as mapping as compared to exact GP regression. Fig. \ref{fig:comp_planners} shows a plot of the progression of the root mean squared error (RMSE) and the mean entropy, calculated on a discretized grid of size $100$x$100$, for the various planning strategies with SSGP regression. Our planning strategy achieves the lowest entropy in the predictions compared to the other methods, which could be attributed to the robot traversing larger distances with our method.

\begin{table*}[h!]
\centering
\begin{tabular}{|c|c|c|c|}
\hline
\textbf{} & \textbf{RMSE} & \textbf{Entropy} & \textbf{Path len. (m)} \\
\hline
Our method ($\epsilon = 0.1$) & $0.18 \pm 0.04$ & \textbf{-1.21 $\pm$ 0.23} & $599.31 \pm 12.08$ \\
\hline
Our method ($\epsilon = 0.2$) & $0.19 \pm 0.04$ & $-1.12 \pm 0.41$ & \textbf{600.57 $\pm$ 13.17} \\
\hline
Our method ($\epsilon = 0.3$) & $0.17 \pm 0.02$ & $-0.93 \pm 0.39$ & $597.37 \pm 9.08$ \\
\hline
RIG-tree Re.h (GP) & $0.20 \pm 0.04$ & $-1.11 \pm 0.16$ & $551.82 \pm 3.34$ \\
\hline
RIG-tree Re.h (SGP) & $0.24 \pm 0.02$ & $-0.06 \pm 0.06$ & $553.45 \pm 28.4$ \\
\hline
RIG-tree Re.h (SSGP) & \textbf{0.14 $\pm$ 0.02} & $-1.16 \pm 0.30$ & $557.89 \pm 1.85$ \\
\hline
Random planner (GP) & $0.28 \pm 0.08$ & $-0.79 \pm 0.21$ & $560.21 \pm 9.54$ \\
\hline
Random planner (SGP) & $0.24 \pm 0.02$ & $0.15 \pm 0.21$ & $564.30 \pm 15.26$ \\
\hline
Random planner (SSGP) & $0.28 \pm 0.05$ & $-0.54 \pm 0.25$ & $555.67 \pm 5.82$ \\
\hline
\end{tabular}
\caption{Comparison of different planning and mapping strategies with the synthetic dataset.}
\label{tab:table1}
\end{table*}


\subsection{Validation with bathymetry dataset}

\begin{figure}
    \centering
    \includegraphics[width=0.47\textwidth]{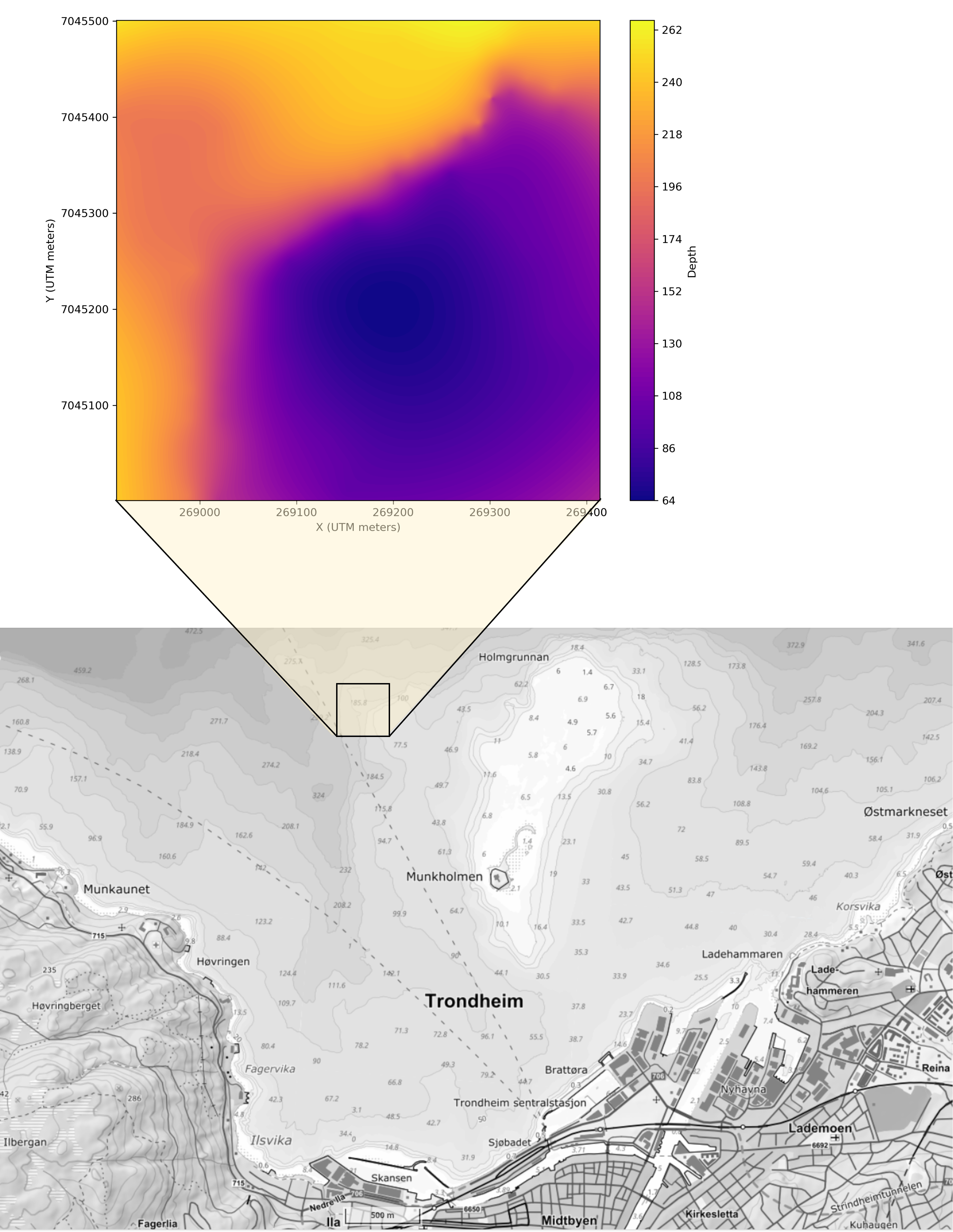}
    \caption{Bathymetry dataset near the coast of Trondheim.}
    \label{fig:map_depth}
\end{figure}

\begin{figure}
    \centering
    \begin{subfigure}{0.48 \textwidth}
        \centering
    \includegraphics[width=0.94\linewidth]{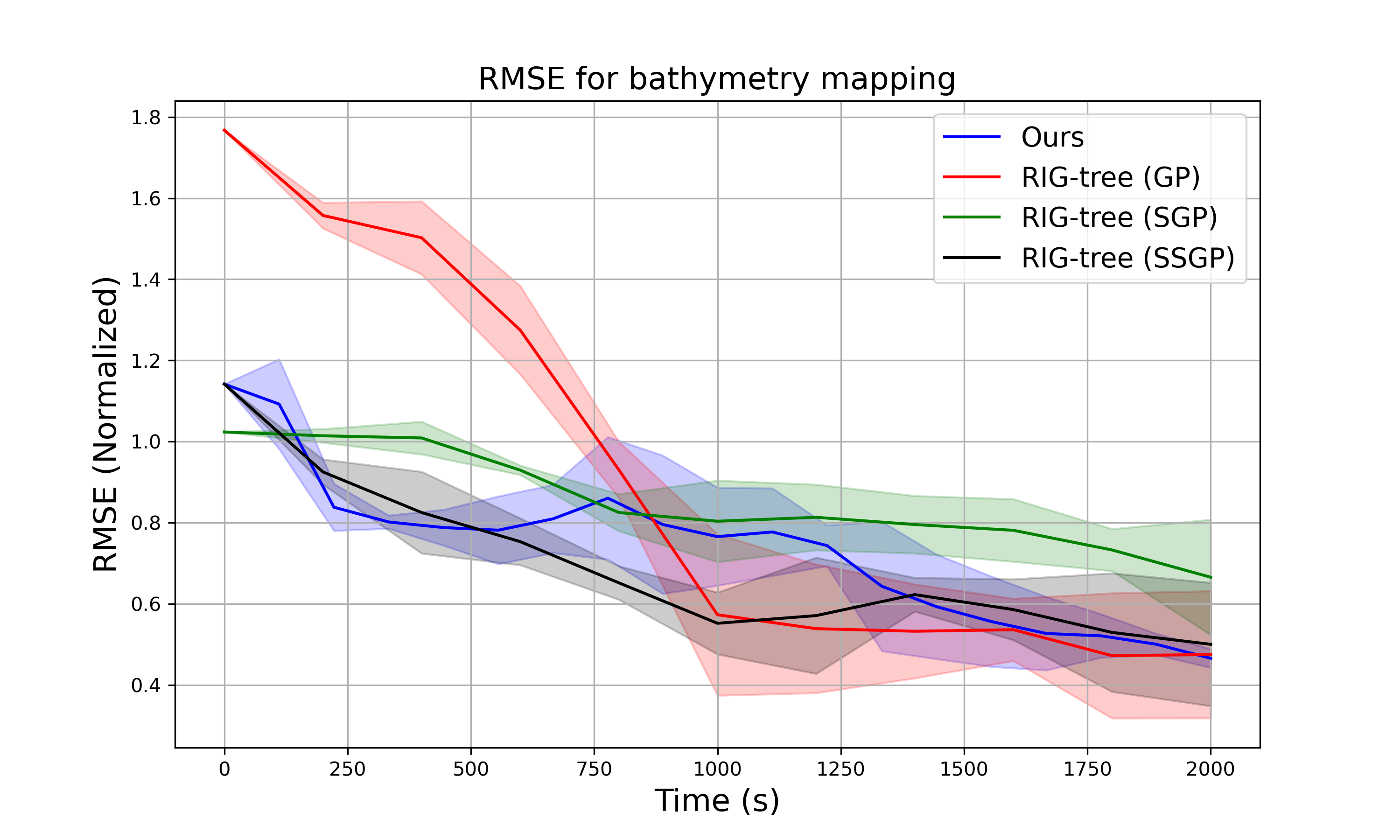}
    \caption{RMSEs for the bathymetry dataset.}
    \label{fig:RMSE_Exp2}
    \end{subfigure}
\begin{subfigure}{0.48 \textwidth}
    \centering
    \includegraphics[width=0.94\linewidth]{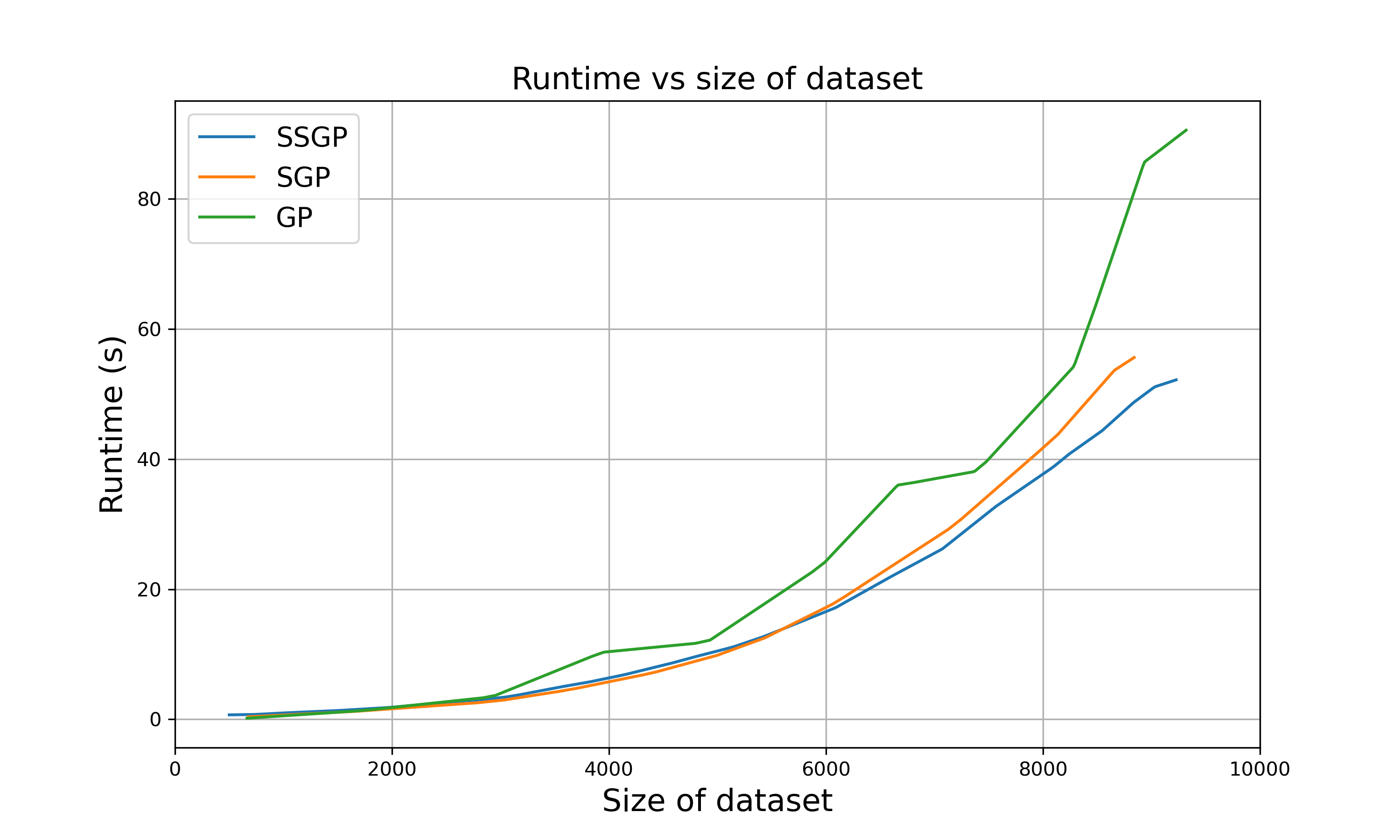}
    \caption{Runtime comparison for the bathymetry dataset.}
    \label{fig:runtime_comp}
\end{subfigure}
\caption{RMSE and GP training runtimes for the bathymetry dataset.}
\label{fig:exp2}
\end{figure}

To evaluate our method for long-term monitoring scenarios, a real dataset with depth measurements in a $500 \times 500$ meter area near the coast of Trondheim is considered, as shown in Fig. \ref{fig:map_depth}. The data is simulated with the VRX-simulator. We set the maximum budget to be $2000$s. Results are reported in Table \ref{tab:tab2}, and visualized in Fig. \ref{fig:exp2}.

\begin{table}[h!]
\centering
\begin{tabular}{|c|c|c|}
\hline
\textbf{} & \textbf{RMSE} & \textbf{Compute time (s)} \\
\hline
Our method & \textbf{0.46} $\pm$ \textbf{0.03} & $340.54 \pm 65.44$ \\
\hline
RIG-tree Re.h (SSGP)& $0.50 \pm 0.15$ &  \textbf{184.10} $\pm$ \textbf{30.23} \\
\hline
RIG-tree Re.h (SGP)& $0.66 \pm 0.14$ &  $199.88 \pm 46.4$ \\
\hline
RIG-tree Re.h (GP)& $0.47 \pm 0.15$ & $279.92 \pm 46.82$ \\
\hline
\end{tabular}
\caption{Results from the bathymetry mapping example.}
\label{tab:tab2}
\end{table}

It can be observed from Fig. \ref{fig:RMSE_Exp2} that our method achieves a marginally better mapping error as that of the RIG-tree method with exact GP and SSGP. Exact GP regression reports better RMSE as compared to the sparse regression techniques with the RIG-tree planner. For this experiment, the total number of samples collected at the end of the run is $\approx10000$. Although the overall computation times, as seen in Table \ref{tab:tab2} between exact GP and SSGP regression are comparable, the latter is updated more frequently and consumes less memory. Fig. \ref{fig:runtime_comp} plots runtimes for the various GP regression techniques against the growth in the size of the dataset. With exact GP regression, runtimes for the hyperparameter optimization scale as $O(n^3)$, and it quickly becomes intractable. Meanwhile, with SSGP regression, the training runtime complexity is $O(bm^2)$. We keep the batch size $b$ constant, and increase the number of inducing points $m$ linearly. Therefore, the training runtime is quadratic in complexity in $n$. Moreover, the memory complexity for exact GP regression is $O(n^2)$, and it is $O(m^2)$ for SVGP and SSGP regression. Both SSGP and SVGP regression are performed with gradient descent in batches, which eases the computational burden. Furthermore, this lower computational burden allows the replanning and the GP training to run in parallel without the vehicle having to stop for replanning.

\section{Conclusions and Future Work}
\label{sec:5}

This paper presented an online information gathering approach for mapping unknown continuous spatial fields. A sampling-based planner is employed for efficient replanning, and streaming sparse GPs for scalable online and incremental learning of the map. Results show that our method achieves similar accuracies to the compared baselines while reducing computation times. Furthermore, the proposed replanning and scalable online learning strategies make it suitable for long-term monitoring scenarios. Future studies shall consider real monitoring tasks to further validate this framework in field experiments. Our method focused on mapping static 2D scalar fields, however, however, its efficacy for mapping spatiotemporal fields or in 3D environments is yet to be investigated. While RRTs easily allow for planning with static obstacles, planning in the presence of dynamic obstacles is not trivial. Therefore, future work could also consider planning in more complex environments containing multiple dynamic obstacles. 

\bibliographystyle{IEEEtranBST/IEEEtran}
\bibliography{refs}

\end{document}